\documentclass[runningheads]{llncs}
\usepackage{graphicx}

\usepackage{tikz}
\usepackage{comment}
\usepackage{amsmath,amssymb} 
\usepackage{color}
\usepackage{hyperref}
\usepackage{multirow}
\usepackage{multicol}
\usepackage{color}
\definecolor{wenxiu}{RGB}{1,1,1}
\newcommand{\wenxiu}{\textcolor{wenxiu}}

\usepackage[accsupp]{axessibility}  

\begin{document}
\pagestyle{headings}
\mainmatter
\def\ECCVSubNumber{8}  

\title{MIPI 2022 Challenge on RGB+ToF Depth Completion: Dataset and Report}

\titlerunning{MIPI 2022 Challenge on RGB+ToF Depth Completion}
%
\author{Wenxiu Sun \and
Qingpeng Zhu \and
Chongyi Li \and
Ruicheng Feng \and
Shangchen Zhou \and
Jun Jiang \and
Qingyu Yang \and
Chen Change Loy \and
Jinwei Gu \and
Dewang Hou \and
Kai Zhao \and
Liying Lu \and
Yu Li \and
Huaijia Lin \and
Ruizheng Wu \and
Jiangbo Lu \and
Jiaya Jia \and
Qiang Liu \and
Haosong Yue \and
Danyang Cao \and
Lehang Yu \and
Jiaxuan Quan \and
Jixiang Liang \and
Yufei Wang \and
Yuchao Dai \and
Peng Yang \and
Hu Yan \and
Houbiao Liu \and
Siyuan Su \and
Xuanhe Li \and
Rui Ren \and
Yunlong Liu \and
Yufan Zhu \and
Dong Lao \and
Alex Wong \and
Katie Chang
\institute{~}
\vspace{-1cm}
}
\authorrunning{W. Sun et al.}
%

\maketitle
\let\thefootnote\relax\footnotetext{\tiny Wenxiu Sun$^{1,3}$ (\email{sunwx@tetras.ai}), Qingpeng Zhu$^{1}$ (\email{zhuqingpeng@tetras.ai}), Chongyi Li$^{4}$, Shangchen Zhou$^{4}$, Ruicheng Feng$^{4}$, Jun Jiang$^{2}$, 
Qingyu Yang$^{2}$, Chen Change Loy$^{4}$, Jinwei Gu$^{2,3}$ are the MIPI 2022 challenge organizers ($^{1}$SenseTime Research and Tetras.AI, $^{2}$SenseBrain, $^{3}$Shanghai AI Laboratory, $^{4}$Nanyang Technological University). The other authors participated in the challenge. Please refer to Appendix~\ref{appendix:teams} for details.\\ 
\\
MIPI 2022 challenge website: \url{http://mipi-challenge.org/}
}
\begin{abstract}
Developing and integrating advanced image sensors with novel algorithms in camera systems is prevalent with the increasing demand for computational photography and imaging on mobile platforms. However, the lack of high-quality data for research and the rare opportunity for in-depth exchange of views from industry and academia constrain the development of mobile intelligent photography and imaging (MIPI). To bridge the gap, we introduce the first MIPI challenge including five tracks focusing on novel image sensors and imaging algorithms. In this paper, RGB+ToF Depth Completion, one of the five tracks, working on the fusion of RGB sensor and ToF sensor (with spot illumination) is introduced. The participants were provided with a new dataset called TetrasRGBD, which contains 18k pairs of high-quality synthetic RGB+Depth training data and 2.3k pairs of testing data from mixed sources. All the data are collected in \wenxiu{an} indoor scenario. We require that the running time of all methods should be real-time \wenxiu{on} desktop GPUs. The final results are evaluated using objective metrics and Mean Opinion Score (MOS) subjectively. A detailed description of all models developed in this challenge is provided in this paper. More details of this challenge and the link to the dataset can be found \wenxiu{at} \href{https://github.com/mipi-challenge/MIPI2022}{https://github.com/mipi-challenge/MIPI2022}

\end{abstract}

\section{Introduction}

RGB+ToF Depth Completion uses sparse ToF depth measurements and a pre-aligned RGB image to obtain a complete depth map. There are \wenxiu{a} few advantages of using sparse ToF depth measurements. First, the hardware power consumption of sparse ToF (with spot illumination) is low compared to full-field ToF (with flood illumination) depth measurements, which is very important for mobile applications to prevent overheat\wenxiu{ing} and fast battery drain. Second, the sparse depth has higher precision and long-range depth measurements due to the focused energy in each laser bin. Third, the Multi-path Interference (MPI) problem that usually bothers full-field iToF measurement is greatly diminished in the sparse ToF depth measurement. However, one obvious disadvantage of sparse ToF depth measurement is the depth density. Take iPhone 12 Pro~\cite{luetzenburg2021evaluation} for example, which is equipped with an advanced mobile Lidar (dToF), the maximum raw depth density is only 1.2\%. If directly use as is, the depth would be too sparse to be applied to typical applications like image enhancement, 3D reconstruction, and AR/VR applications.

In this challenge, we intend to fuse the pre-aligned RGB and sparse ToF depth measurement to obtain a complete depth map. Since the power consumption and processing time are still  important factors to be balanced, in this challenge, the proposed algorithm is required to process the RGBD data and predict the depth in real-time, i.e. reaches speeds of 30 frames per second, on a reference platform with a GeForce RTX 2080 Ti GPU. The solution is not necessarily deep learning solution, however, to facilitate the deep learning training, we provide a high-quality synthetic depth training dataset containing 20,000 pairs of RGB and ground truth depth images of 7 indoor scenes. We provide a data loader to read these files and a function to simulate sparse depth maps that \wenxiu{are} close to the real sensor measurements. The participants are also allowed to use other public-domain dataset, for example, NYU-Depth v2~\cite{silberman2012indoor}, KITTI Depth Completion dataset~\cite{Uhrig2017THREEDV},  Scenenet~\cite{mccormac2017scenenet}, Arkitscenes~\cite{baruch2021arkitscenes}, Waymo open dataset~\cite{sun2020scalability}, etc. A baseline code is available as well to understand the whole pipeline and to wrap up quickly. The testing data comes from mixed sources, including synthetic data, spot-iToF, and samples that are manually subsampled from iPhone 12 Pro processed depth which uses dToF. The algorithm performance will be ranked by objective metrics: relative mean absolute error (RMAE), edge-weighted mean absolute error (EWMAE), relative depth shift (RDS), and relative temporal standard deviation (RTSD). Details of the metrics are described in the Evaluation section. For \wenxiu{the} final evaluation, we will also evaluate subjectively in metrics that could not be measured effectively using objective metrics, for example, XY-resolution, edge sharpness, smoothness on flat surfaces, etc.

This challenge is a part of the Mobile Intelligent Photography and Imaging (MIPI) 2022 workshop and challenges which emphasize the integration of novel image sensors and imaging algorithms, which is held in conjunction with ECCV 2022. It consists of five competition tracks:
\begin{enumerate}
  \item RGB+ToF Depth Completion  uses sparse, noisy ToF depth measurements with RGB images to obtain a complete depth map.
  \item Quad-Bayer Re-mosaic  converts Quad-Bayer RAW data into Bayer format so that it can be processed with standard ISPs.
  \item RGBW Sensor Re-mosaic  converts RGBW RAW data into Bayer format so that it can be processed with standard ISPs.
  \item RGBW Sensor Fusion  fuses Bayer data and a monochrome channel data into Bayer format to increase SNR and spatial resolution.
  \item Under-display Camera Image Restoration  improves the visual quality of image captured by a new imaging system equipped with under-display camera.
\end{enumerate}

\section{Challenge}
To develop \wenxiu{an} efficient and high-performance RGB+ToF Depth Completion solution to be used for mobile applications, we provide the following resources for participants:
\begin{itemize}
    \item A high-quality and large-scale dataset that can be used to train and test the solution;
    \item The data processing code with data loader that can help participants to save time to accommodate the provided dataset to the depth completion task;
    \item A set of evaluation metrics that can measure the performance of \wenxiu{a} developed solution;
    \item A suggested requirement of running time on multiple platforms that is necessary for real-time processing.
\end{itemize}

\subsection{Problem Definition}
Depth completion~\cite{hu2022deep,eldesokey2020uncertainty,eldesokey2019confidence,hu2021penet,imran2019depth,li2020multi,lopez2020project,qu2020depth,chen2018estimating,cheng2019learning,ma2019self,park2020non,lee2021depth} aims to recover dense depth from sparse depth measurements. Earlier methods concentrate on retrieving dense depth maps only from the sparse ones. However, these approaches are limited and not able to recover depth details and semantic information without the availability of multi-modal data.
In this challenge, we focus on the RGB+ToF sensor fusion, where a pre-aligned RGB image is also available as guidance for depth completion. 

In our evaluation, the depth resolution and RGB resolution are fixed at $256 \times 192$, and the input depth map sparsity ranges from $1.0\%$ to $1.8\%$. As a reference, the KITTI depth completion dataset~\cite{Uhrig2017THREEDV} has sparsity around $4\% \sim 5\%$. The target of this challenge is to predict a dense depth map given the sparsity
depth map and a pre-aligned RGB image at \wenxiu{the} allowed running time constraint (please refer to Section~\ref{sec:runtime} for details).

\subsection{Dataset: TetrasRGBD}

\begin{figure}[!ht]
\centering
\includegraphics[width=\textwidth]{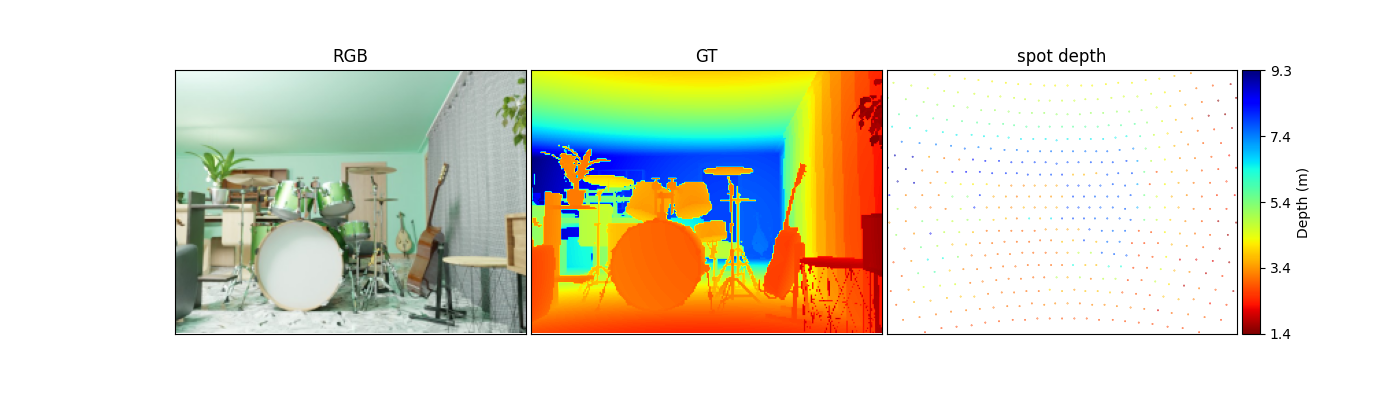}
\caption{Visualization of an example training data. Left, middle, and right images correspond to RGB, ground-truth depth, and simulated spot depth data, respectively.}
\label{fig:sample}
\setlength{\belowcaptionskip}{0pt plus 3pt minus 2pt}
\end{figure}

The training data contains 7 image sequences of aligned RGB and ground-truth dense depth from 7 indoor scenes (20,000 pairs of RGB and depth in total). For each scene, the RGB and the ground-truth depth are rendered along a smooth trajectory in our created 3D virtual environment. RGB and dense depth images in the training set have a resolution of 640$\times$480 pixels. We also provide a function to simulate the sparse depth maps that are close to the real sensor measurements\footnote{https://github.com/zhuqingpeng/MIPI2022-RGB-ToF-depth-completion}. A visualization of an example frame of RGB, ground-truth depth, and simulated sparse depth is shown in Fig.~\ref{fig:sample}.  

The testing data contains, a) \textit{Synthetic}: a synthetic image sequence (500 pairs of RGB and depth in total) rendered from an indoor virtual environment that differs from the training data; b) \textit{iPhone dynamic}: 24 image sequences of dynamic scenes collected from an iPhone 12Pro (600 pairs of RGB and depth in total); c) \textit{iPhone static}: 24 image sequences of static scenes collected from an iPhone 12Pro (600 pairs of RGB and depth in total);  d) \textit{Modified phone static}: 24 image sequences of static scenes (600 pairs of RGB and depth in total) collected from a modified phone. Please note that depth noises, missing depth values in low reflectance regions, and mismatch of field of views between RGB and ToF cameras could be observed from this real data. RGB and dense depth images in the entire testing set have the resolution of 256$\times$192 pixels. RGB and spot depth data from the testing set are provided and the GT depth are not available to participants. The depth data in both training and testing sets are in meters.

\subsection{Challenge Phases}
The challenge consisted of the following phases:
\begin{enumerate}
    \item Development: The registered participants get access to the data and baseline code, and are able to train the models and evaluate their running time locally.
    \item Validation: The participants can upload their models to the remote server to check the fidelity scores on the validation dataset, and to compare their results on the validation leaderboard.
    \item Testing: The participants submit their final results, code, models, and factsheets.
\end{enumerate}

\subsection{Scoring System}
\subsubsection{Objective Evaluation}
We define the following metrics to evaluate the performance of depth completion algorithms.
\begin{itemize}
    \item Relative Mean Absolute Error (RMAE), which measures the relative depth error between the completed depth and the ground truth, i.e.
    \begin{equation}
    \small
    \text{RMAE} = \frac{1}{M \cdot N}\sum_{m=1}^{M}\sum_{n=1}^{N} \left\lvert\frac{\hat{D}(m,n)-D(m,n)}{D(m,n)}\right\rvert,
    \end{equation}
    where $M$ and $N$ denote the height and width of depth, respectively. $D$ and $\hat{D}$ represent the ground-truth depth and the predicted depth, respectively.

    \item Edge Weighted Mean Absolute Error (EWMAE), which is a weighted average of absolute error. Regions with larger depth discontinuity are assigned higher weights. Similar to the idea of Gradient Conduction Mean Square Error (GCMSE) \cite{lopez2017}, EWMAE applies a weighting coefficient $G(m,n)$ to the absolute error between pixel $(m,n)$ in ground-truth depth $D$ and predicted depth $\hat{D}$, i.e.
    \begin{equation}
    \small
    \text{EWMAE} = \frac{1}{M \cdot N} \left\lvert\frac{\sum_{m=1}^{M}\sum_{n=1}^{N}G(m,n)\cdot[\hat{D}(m,n)-D(m,n)]}{\sum_{m=1}^{M}\sum_{n=1}^{N}G(m,n)}\right\rvert,
    \end{equation}
    where the weight coefficient $G$ is computed in the same way as in \cite{lopez2017}.
    
    \item Relative Depth Shift (RDS), which measures the relative depth error between the completed depth and the input sparse depth on the set of pixels where there are valid depth values in the input sparse depth, i.e.
    \begin{equation}
    \small
    \text{RDS} = \frac{1}{n(S)}\sum_{s\in S} \left\lvert\frac{\hat{D}(s)-X(s)}{X(s)}\right\rvert,
    \end{equation}
    where $X$ represents the input sparse depth. $S$ denotes the set of the coordinates of all spot pixels, i.e. pixels with valid depth value. $n(S)$ denotes the cardinality of $S$.

    \item Relative Temporal Standard Deviation (RTSD), which measures the temporal standard deviation normalized by depth values for static scenes, i.e.
    \begin{equation}
    \small
    \text{RTSD} =  \frac{1}{M \cdot N}\left( \sqrt{\frac{\sum_{f=1}^{F}\left(\hat{D}(m,n)-\sum_{f=1}^{F}\hat{D}(m,n)\right)}{F}} \bigg /\sum_{f=1}^{F}\hat{D}(m,n)\right),
    \end{equation}    
    where $F$ denotes the number of frames.
    
\end{itemize}

RMAE, EWMAE, and RDS will be measured on the testing data with GT depth. RTSD will be measured on the testing data collected from static scenes.
We will rank the proposed algorithms according to the score calculated by the following formula, where the coefficients are designed to balance the values of different metrics,
\begin{equation}
    \text{Objective score} = 1 - 1.8 \times \text{RMAE} - 0.6 \times \text{EWMAE} - 3 \times \text{RDS} - 4.6 \times \text{RTSD}.
\end{equation}
For each dataset, we report the average results over all the processed images belonging to it.

\subsubsection{Subjective Evaluation}
For subjective evaluation, we adapt the commonly used Mean Opinion Score (MOS) with blind evaluation. The score is on a scale of 1 (bad) to 5 (excellent). We invited 16 expert observers to watch videos and give their subjective score independently. The scores of all subjects are averaged as the final MOS.

\subsection{Running Time Evaluation}
\label{sec:runtime}
The proposed algorithms are required to be able to process the RGB and sparse depth sequence  in real-time.  Participants are required to include the average run time of one pair of RGB and depth data using their algorithms and the information \wenxiu{in} the device in the submitted readme file. Due to the difference of devices for evaluation, we set different requirements of running time for different types of devices according to the AI benchmark data from the website\footnote{https://ai-benchmark.com/ranking\_deeplearning\_detailed.html}.  Although the running time is still far from real-time and low power consumption
on edge computing devices, we believe this could set up a good starting point for researchers to further push the limit in both academia and industry.

\section{Challenge Results}

From $119$ registered participants, $18$ teams submitted their results in the validation phase, $9$ teams entered the final phase and submitted the valid results, code, executables, and factsheets. Table~\ref{tab:results} summarizes the final challenge results. Team 5 (ZoomNeXt) shows the best overall performance, followed by Team 1 (GAMEON) and Team 4 (Singer). The proposed methods are described in Section \ref{sec:methods} and the team members and affiliations are listed in Appendix \ref{appendix:teams}.

\begin{table}[!ht]  
    \centering
    \tiny
    \begin{tabular}{c|l|llll|l|l|l}
    \hline
        \textbf{Team No.} & \textbf{Team name/User name} & \textbf{RMAE} & \textbf{EWMAE} & \textbf{RDS} & \textbf{RTSD} & \textbf{Objective} & \textbf{Subjective} & \textbf{Final}\\ \hline  \hline
        5 & ZoomNeXt/Devonn, k-zha14 & 0.02935  & 0.13928  & 0.00004  & 0.00997  & 0.81763  & 3.53125  & \textbf{0.76194}  \\ \hline
        1 & GAMEON/hail\_hydra & \textbf{0.02183}  & \textbf{0.13256}  & 0.00736  & 0.01127  & 0.80723  & \textbf{3.55469}  & 0.75908  \\ \hline
        4 & Singer/Yaxiong\_Liu & 0.02651  & 0.13843  & 0.00270  & 0.01012  & 0.81457  & 3.32812  & 0.74010  \\ \hline
        0 & NPU-CVR/Arya22 & 0.02497  & 0.13278  & 0.00011  & 0.00747  & \textbf{0.84071}  & 3.14844  & 0.73520  \\ \hline
        2 & JingAM/JingAM & 0.03767  & 0.13826  & 0.00459  & 0.00000  & 0.83545  & 2.96094  & 0.71382  \\ \hline
        6 & NPU-CVR/jokerWRN & 0.02547  & 0.13418  & 0.00101  & 0.00725  & 0.83729  & 2.75781  & 0.69443  \\ \hline
        8 & Anonymous/anonymous & 0.03015  & 0.13627  & 0.00002  & 0.01716  & 0.78497  & 2.76562  & 0.66905  \\ \hline
        3 & MainHouse113/renruixdu & 0.03167  & 0.14771  & 0.01103  & 0.01162  & 0.76781  & 2.71875  & 0.65578  \\ \hline
        7 & UCLA Vision Lab/laid & 0.03890  & 0.14731  & 0.00014  & 0.00028  & 0.83990  & 2.09375  & 0.62933  \\ \hline
    \end{tabular}
    \caption{MIPI 2022 RGB+ToF Depth Completion challenge results and final rankings. Team ZoomNeXt is the challenge winner.
    \label{tab:results}}
\end{table}

To analyze the performance on different testing dataset, we also summarized the objective score (RMAE) and the subjective score (MOS) per dataset in Table~\ref{tab:score-per-dataset}, namely \textit{Synthetic}, \textit{iPhone dynamic}, \textit{iPhone static}, and \textit{Modified phone static}. Note that due to there is no RTSD score for dynamic datasets and the RDS in the submitted results \wenxiu{is} usually very small if participants used hard depth replacement, we only present the RMAE score in this table as an objective indicator. Team ZoomNeXt performs the best in the \textit{Modified phone static} subset, and moderate in other subsets. Team GAMEON performs the best in the subset of \textit{Synthetic}, \textit{iPhone dynamic}, and \textit{iPhone static}, however, obvious artifacts could be observed in the \textit{Modified phone static} subset.

\begin{table}[!ht]
\setlength{\abovecaptionskip}{0.cm}
\setlength{\belowcaptionskip}{-0.cm}
    \centering
    \tiny
    \begin{tabular}{c|l|l|l|l|l|l|l|c|l}
    \hline
        \multirow{2}{*}{\textbf{Team No.}}  & \multirow{2}{*}{\textbf{Team Name/User Name}} & \multicolumn{2}{|l|}{\textbf{Synthetic}} & \multicolumn{2}{|l|}{\textbf{iPhone dynamic}}   & \multicolumn{2}{|l|}{\textbf{iPhone static}}  & \multicolumn{2}{|l}{\textbf{Mod. phone static}}  \\ 
        ~ & ~ & RMAE & MOS & RMAE & MOS & RMAE & MOS & RMAE & MOS  \\ \hline
        5  & ZoomNeXt/Devonn, k-zha14 & 0.06478  & 3.37500 & 0.01462  & 3.53125  & 0.01455  & 3.40625  & / & \textbf{3.8125}     \\ \hline
        1  & GAMEON/hail\_hydra  & \textbf{0.05222}  & \textbf{4.18750} & \textbf{0.00919}  & \textbf{3.96875}  & \textbf{0.00915}  & \textbf{3.84375}  & / & 2.21875   \\ \hline
        4  & Singer/Yaxiong\_Liu  & 0.06112  & 3.34375 & 0.01264  & 3.68750  & 0.01154  & 2.71875  & / & 3.56250     \\ \hline
        0  & NPU-CVR/Arya22  & 0.05940  & 3.50000 & 0.01099  & 3.09375  & 0.01026  & 3.25000  & / & 2.75000     \\ \hline
        2  & JingAM/JingAM  & 0.0854  & 2.50000 & 0.01685  & 3.50000  & 0.01872  & 3.28125  & / & 2.56250     \\ \hline
        6  & NPU-CVR/jokerWRN  & 0.06061  & 3.15625 & 0.01116  & 2.62500  & 0.01050  & 2.87500  & / & 2.37500    \\ \hline
        8  & Anonymous/anonymous  & 0.06458  & 3.87500 & 0.01589  & 2.75000  & 0.01571  & 2.53125  & / & 1.90625     \\ \hline
        3  & MainHouse113/renruixdu  & 0.06928  & 2.90625 & 0.01718  & 3.06250  & 0.01482  & 2.03125  & / & 2.87500    \\ \hline
        7 & UCLA Vision Lab/laid & 0.08677 & 1.71875 & 0.01997 & 2.18750 & 0.01793 & 2.12500 & / & 2.34375   \\ \hline
    \end{tabular}
    \caption{Evalutions of RMAE and MOS in the testing dataset.
    \label{tab:score-per-dataset}}
\end{table}

\begin{figure}[!ht]
\setlength{\abovecaptionskip}{0.cm}
\setlength{\belowcaptionskip}{-0.cm}
\centering
\includegraphics[width=0.45\textwidth]{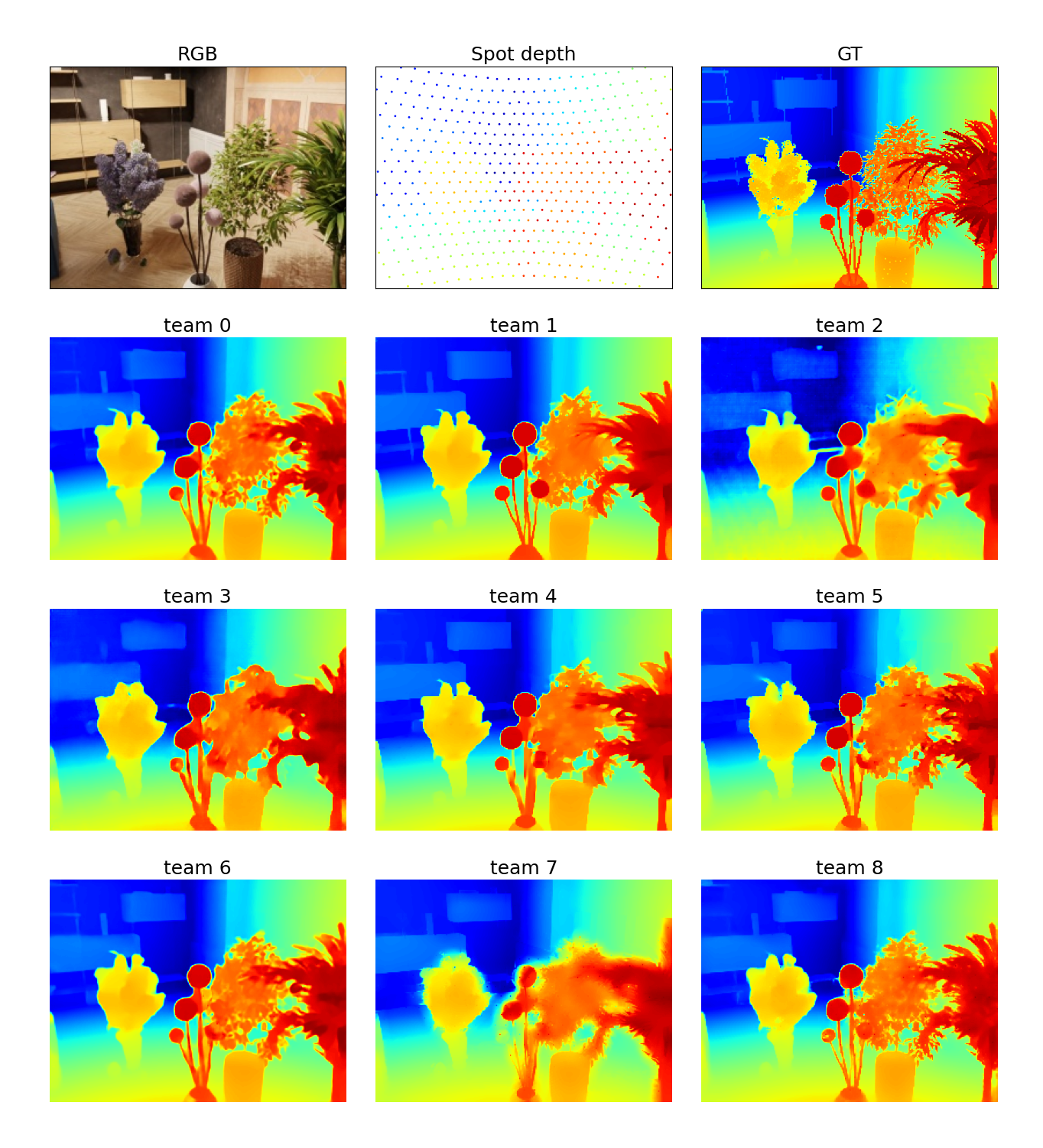}
\includegraphics[width=0.45\textwidth]{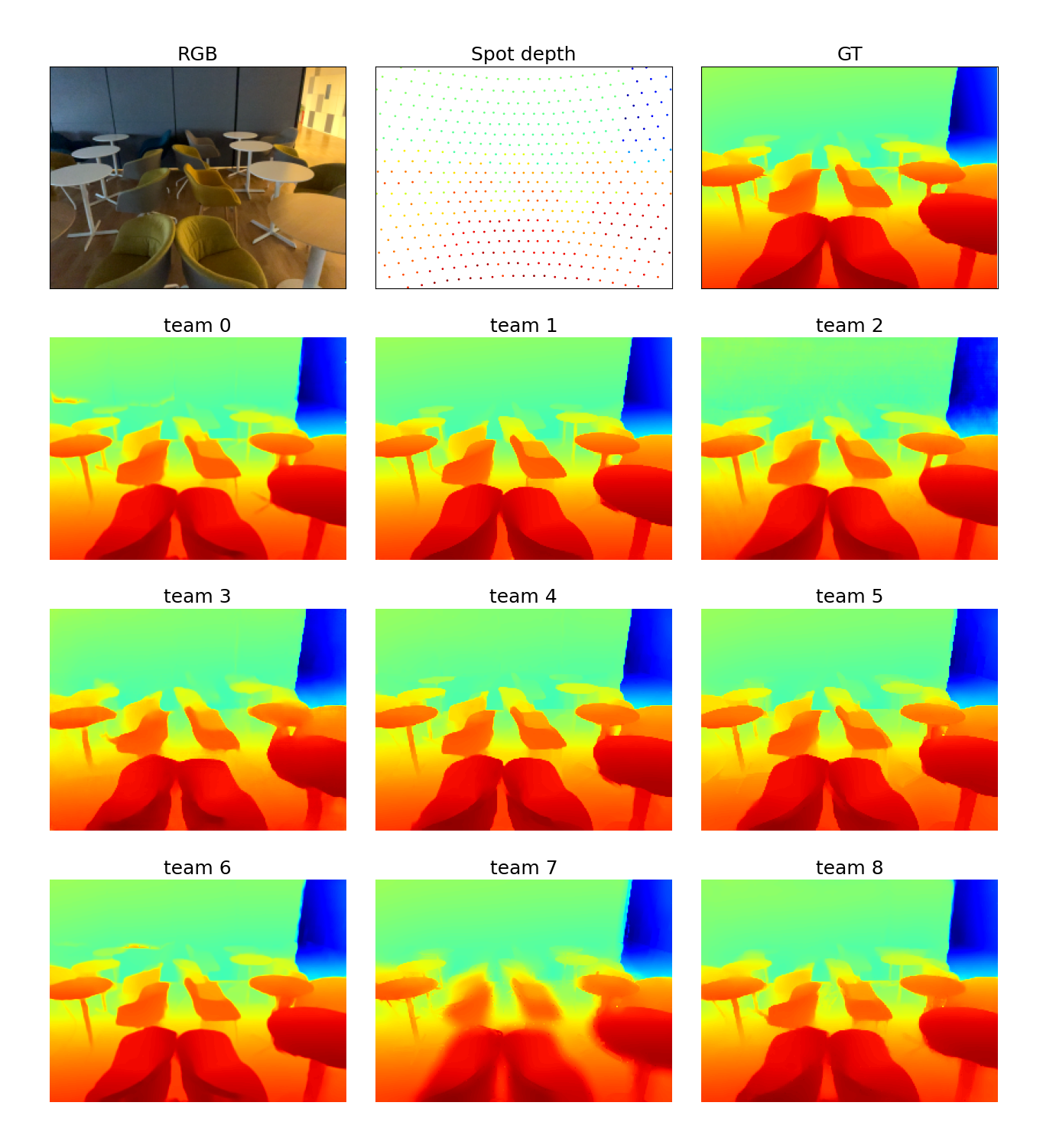}
\includegraphics[width=0.45\textwidth]{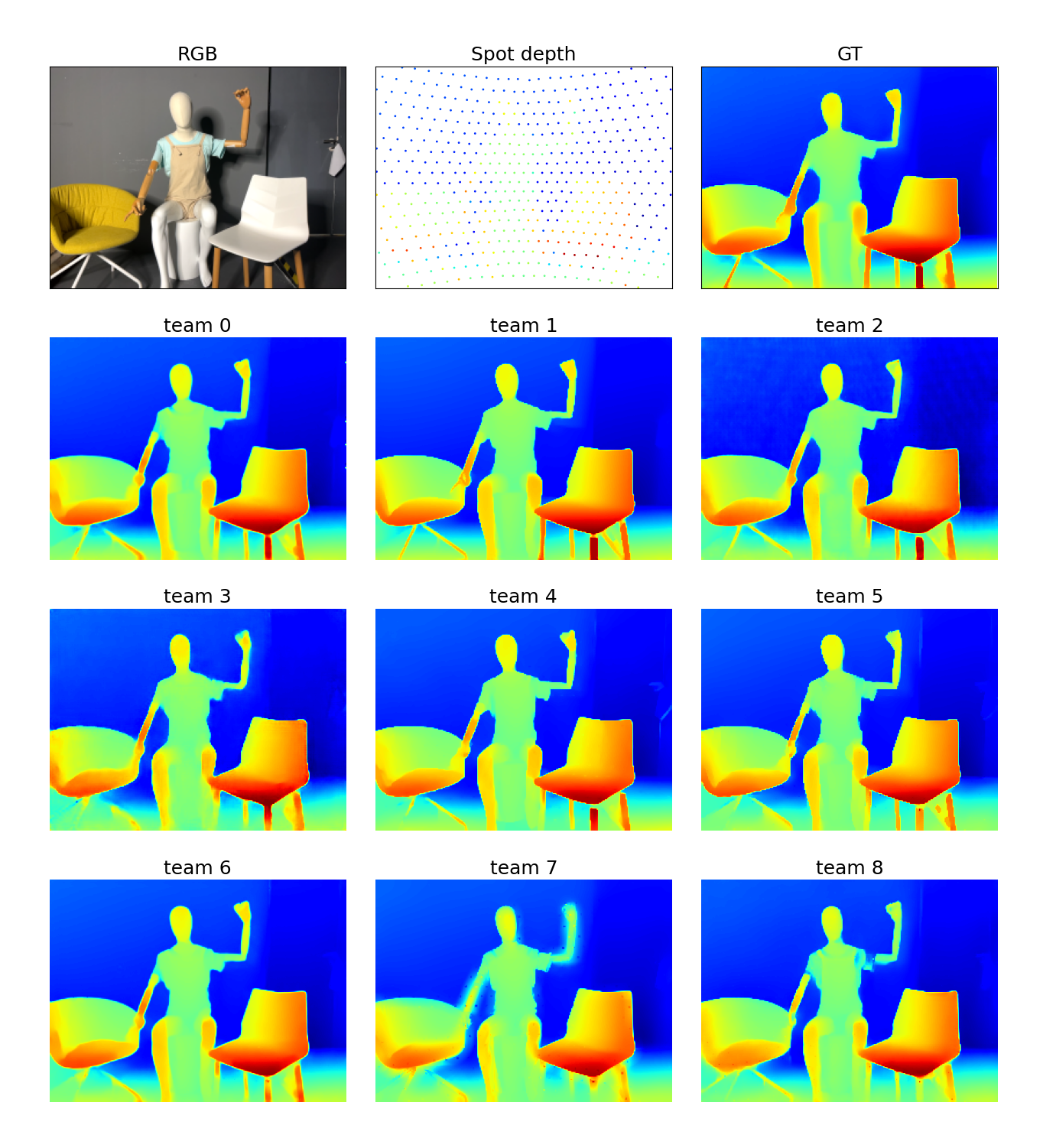}
\includegraphics[width=0.45\textwidth]{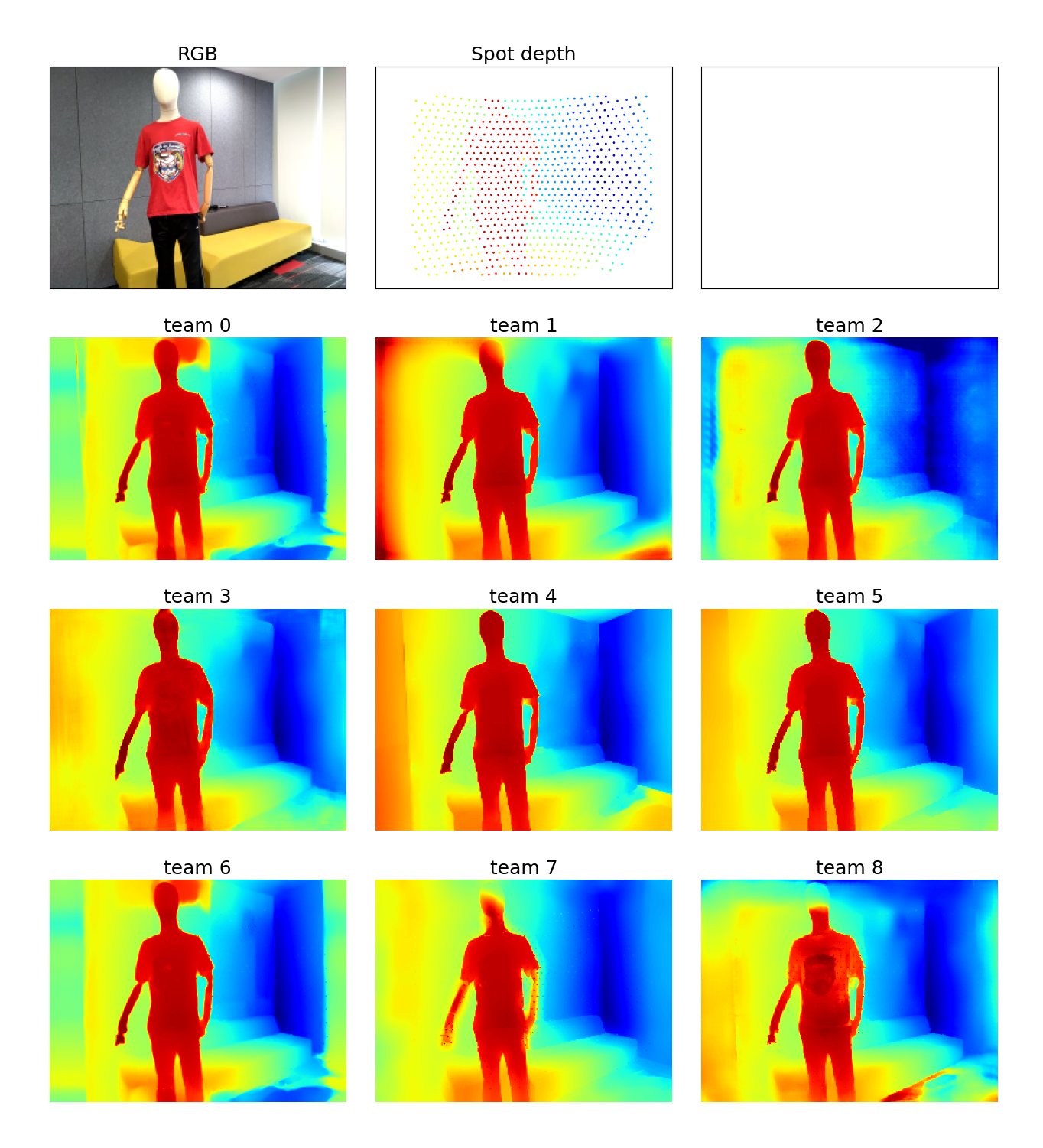}
\caption{\wenxiu{Visualization} of results. The set of images shown in the top left \wenxiu{is} from \wenxiu{the} Synthetic subset including RGB image, sparse depth, ground-truth depth (if available), and results of 9 teams. Similarly, the set of images shown in \wenxiu{the} top right \wenxiu{is} from iPhone dynamic subset, the set of images shown in \wenxiu{the} bottom left \wenxiu{is} from \wenxiu{the} iPhone static subset, and the set of images shown in \wenxiu{the} bottom right \wenxiu{is} from \wenxiu{the} Modified phone subset.
}
\label{fig:results_vis}
\end{figure}

Fig.~\ref{fig:results_vis} shows a single frame visualization of all the submitted results in the test dataset. Top-left pair is from \textit{Synthetic} subset, top-right pair is from \textit{iPhone dynamic} subset, bottom-left pair is from \textit{iPhone static} subset, and bottom-right pair is from \textit{Modified phone static} subset. It can be observed that all the models could reproduce a semantically meaningful dense depth map. Team 1 (GAMEON) shows the best XY-resolution, where the tiny structures (chair legs, fingers, etc.) could be correctly estimated. Team 5 (ZoomNeXt) and Team 4 (Singer) \wenxiu{show} \wenxiu{the} most stable cross-dataset performance, especially in the modified phone dataset.

The running time of submitted methods on their individual platforms is summarized in Table~\ref{tab:runtime}. All methods satisfied the real-time requirements when converted to a reference platform with a GeForce RTX 2080 Ti GPU. 

\begin{table}[!ht]
    \centering
    \tiny
    \begin{tabular}{c|l|c|l|c}
    \hline
        \textbf{Team No.} & \textbf{Team name/User name} & \textbf{Inference Time} & \textbf{Testing Platform} & \textbf{Upper Limit} \\ \hline
        5 & ZoomNeXt/Devonn, k-zha14 & 18ms & Tesla V100 GPU & 34ms \\ \hline
        1 & GAMEON/hail\_hydra & 33ms & GeForce RTX 2080 Ti & 33ms \\ \hline
        4 & Singer/Yaxiong\_Liu & 33ms & GeForce RTX 2060 SUPER & 50ms \\ \hline
        0 & NPU-CVR/Arya22 & 23ms & GeForce RTX 2080 Ti & 33ms \\ \hline
        2 & JingAM/JingAM & 10ms & GeForce RTX 3090 & / \\ \hline
        6 & NPU-CVR/jokerWRN & 24ms & GeForce RTX 2080 Ti & 33ms \\ \hline
        8 & Anonymous/anonymous & 7ms & GeForce RTX 2080 Ti & 33ms \\ \hline
        3 & MainHouse113/renruixdu & 28ms & GeForce 1080 Ti & 48ms \\ \hline
        7 & UCLA Vision Lab/laid & 81ms & GeForce 1080 Max-Q & 106ms \\ \hline
    \end{tabular}
    \caption{Running time of submitted methods. All of the methods satisfied the real-time inference requirement.
    \label{tab:runtime}}
\end{table}
\vspace{-0.5cm}

\section{Challenge Methods}
\label{sec:methods}
In this section, we describe the solutions submitted by all teams participating in the final stage of MIPI 2022 RGB+ToF Depth Completion Challenge. A brief taxonomy of all the methods is in Table~\ref{tab:methods}. 
\begin{table}[!ht]
    \centering
    \tiny
    \label{tab:methods}
    \begin{tabular}{l|l|l|l|l|l|l}
    \hline
        \textbf{Team name} & \textbf{Fusion} & \textbf{Multi-scale} & \textbf{Refinement} & \textbf{Inspired from} & \textbf{Ensemble} & \textbf{Additional data}\\ \hline
        ZoomNeXt & Late & No & SPN series & FusionNet~\cite{van2019sparse} & No & No \\ \hline
        GAMEON & Early & Yes & SPN series & NLSPN~\cite{park2020non} & No & ARKitScenes~\cite{baruch2021arkitscenes}\\ \hline
        Singer & Late & Yes & SPN series & Sehlnet~\cite{liu2022sehlnet} & No & No \\ \hline
        NPU-CVR & Mid & No & Deformable Conv & GuideNet~\cite{tang2020learning} & No & No \\ \hline
        JingAM & Late & No & No & FusionNet~\cite{van2019sparse} & No & No \\ \hline
        Anonymous & Early & No & No & FCN & Yes (Self) & No\\ \hline
        MainHouse113 & Early & Yes & No & MobileNet~\cite{howard2017mobilenets} & No & No\\ \hline
        UCLA Vision Lab & Late & No & Bilateral Filter & ScaffNet~\cite{wong2021learning}, FusionNet~\cite{van2019sparse} & Yes (Network) & SceneNet~\cite{mccormac2017scenenet}\\ \hline
    \end{tabular}
    \caption{A taxonomy of the all the methods in the final stage.}
\end{table}
\vspace{-0.5cm}

\subsection{ZoomNeXt}

\begin{figure}[!ht]
\setlength{\abovecaptionskip}{0.cm}
\setlength{\belowcaptionskip}{-0.cm}
\centering
\includegraphics[width=0.8\textwidth]{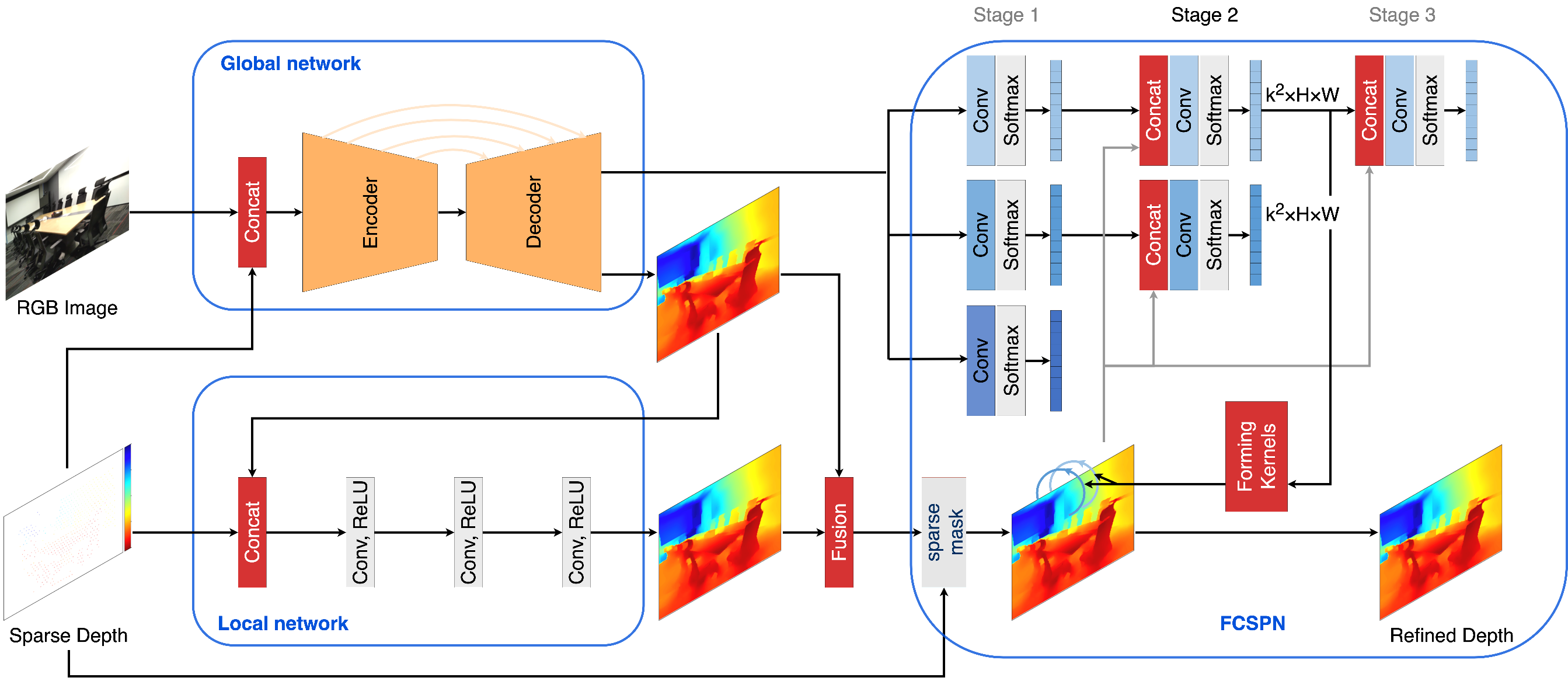}
\caption{Model architecture of ZoomNeXt team.}
\setlength{\belowcaptionskip}{0pt plus 3pt minus 2pt}
\label{fig:zoomnext-network}
\end{figure}
Team ZoomNeXt proposes a lightweight and efficient multimodal depth completion (EMDC) model, shown in Fig.~\ref{fig:zoomnext-network}, which carries their solution to better address the following three key problems.
\begin{enumerate}
    \item How to fuse multi-modality data more effectively? For this problem, the team adopted a Global and Local Depth Prediction (GLDP) framework~\cite{hu2022deep,van2019sparse}. For the confidence maps used by the fusion module, they adjusted the pathways of the features to have relative certainty of global and local depth predictions. Furthermore, the losses calculated on the global and local predictions are adaptively weighted to avoid model mismatch. 
    \item How to reduce the negative effects of missing values regions in sparse modality? They first replaced the traditional upsampling in the U-Net in global network with pixel-shuffle~\cite{shi2016real}, and also removed the batch normalization in local network, as they are fragile to features with anisotropic distribution and degrade the results.
    \item How to better recover scene structures for both objective metrics and subjective quality? They proposed key innovations in terms of SPN~\cite{cheng2018depth,cheng2020cspn++,park2020non,lin2022dynamic,hu2021penet} structure and loss function design. First, they proposed the funnel convolutional spatial propagation network (FCSPN) for depth refinement. FCSPN can fuse the point-wise results from large to small dilated convolutions in each stage, and the maximum dilation at each stage is designed to be gradually smaller, thus forming a funnel-like structure stage by stage. Second, they also proposed a corrected gradient loss to handle the extreme depth (0 or \textit{inf}) in the ground-truth.
\end{enumerate}

\subsection{GAMEON}
\setlength{\abovecaptionskip}{0.cm}
\setlength{\belowcaptionskip}{-0.cm}
\begin{figure}[!ht]
\centering
\includegraphics[width=0.34\textwidth]{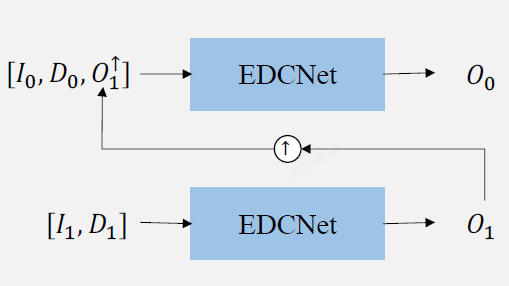}
\includegraphics[width=0.43\textwidth]{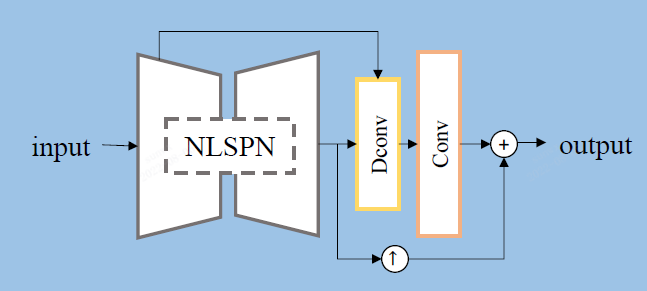}
\caption{Model architecture of GAMEON team.}
\label{fig:gameon-network}
\end{figure}
GAMEON team proposed a multi-scale architecture to complete the sparse depth map with high performance and
fast speed. Knowledge distillation method is used to distill information from large models. Besides,
they proposed a stability constraint to increase the model stability and robustness.

They adopt NLSPN~\cite{park2020non} as their baseline, and improve it in the following several aspects. As shown in the left figure of
Fig.~\ref{fig:gameon-network}, they adopted a multi-scale scheme to produce the final result. At each scale, the proposed
EDCNet is used to generate the dense depth map at the current resolution. The detailed architecture of
EDCNet is shown in the right figure of Fig.~\ref{fig:gameon-network}, where NLSPN is adopted as the base network to produce the dense depth map at 1/2 resolution of the input, then two convolution layers are used to refine and generate
the full resolution result. To enhance the performance, they further adopt knowledge distillation to
distill information from Midas (DPT-Large)~\cite{ranftl2020towards}. They distill information from the penultimate layer of
Midas to the penultimate layer of the depth branch of NLSPN. A convolution layer of kernel size $1$
is used to solve the channel dimension gap. Furthermore, they also propose a stability constraint to
increase the model stability and robustness. In particular, given a training sample, they adopt the thin
plate splines (TPS)~\cite{duchon1977splines} transformation (very small perturbations) on the input and ground-truth to
generate the transformed sample. After getting the outputs for these two samples, they apply the same transformation \wenxiu{to} the output of the original sample, which is constrained to be the same as the output
of the transformed sample.

\subsection{Singer}
\begin{figure}[!ht]
\setlength{\abovecaptionskip}{0.cm}
\setlength{\belowcaptionskip}{-0.cm}
\centering
\includegraphics[width=0.9\textwidth]{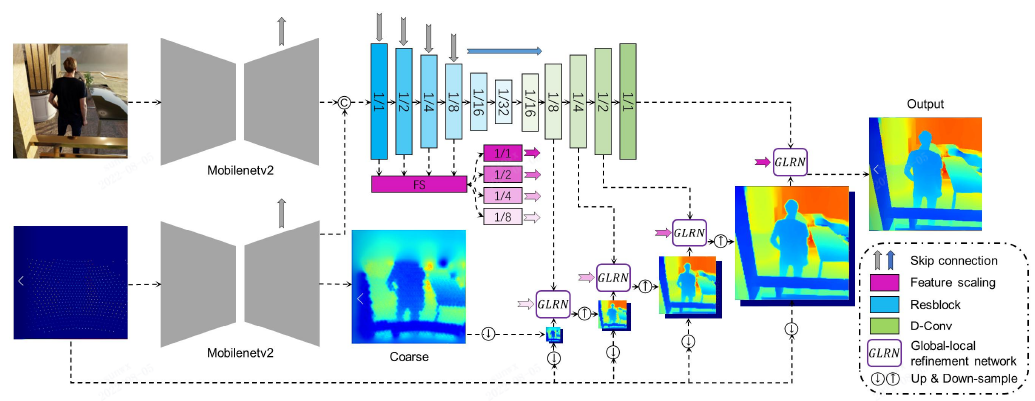}
\caption{Model architecture of Singer team.}
\label{fig:singer-network}
\end{figure}
Team Singer proposed a depth completion approach that separately estimates high- and low-frequency components to address the problem of how to sufficiently utilize of multimodal data. Based on their previous work~\cite{liu2022sehlnet}, they
proposed a novel Laplacian pyramid-based~\cite{song2021monocular,chen2020laplacian,jeon2018reconstruction} depth completion network, which estimates low-frequency components from sparse depth maps by downsampling and contains a Laplacian pyramid decoder that estimates multi-scale residuals to reconstruct complex details of the scene. 
The overall architecture of the network is shown in Fig.~\ref{fig:singer-network}. 

They use two independent mobilenetv2 to
extract features from RGB and Sparse ToF depth. The features from two decoders are fused in a resnet-based encoder. To recover high-frequency scene structures while saving computational cost, the proposed Laplacian pyramid representation progressively adds information on different frequency bands so that the scene structure at different scales can be hierarchically recovered during reconstruction. 
Instead of the simple upsampling and summation of the two parts in two
frequency bands, they proposed a global-local refinement network (GLRN) to fully use different levels of features
from the encoder to estimate residuals at various scales and refine them. 
To save the computational cost of using spatial propagation networks, 
they introduced a dynamic mask for the fixed kernel of CSPN termed as \wenxiu{Affinity} Decay Spatial Propagation Network (AD-SPN), which is used to refine the estimated depth maps at
various scales through spatial propagation.

\subsection{NPU-CVR}
\begin{figure}[!ht]
\setlength{\abovecaptionskip}{0.cm}
\setlength{\belowcaptionskip}{-0.cm}
\centering
\includegraphics[width=0.7\textwidth]{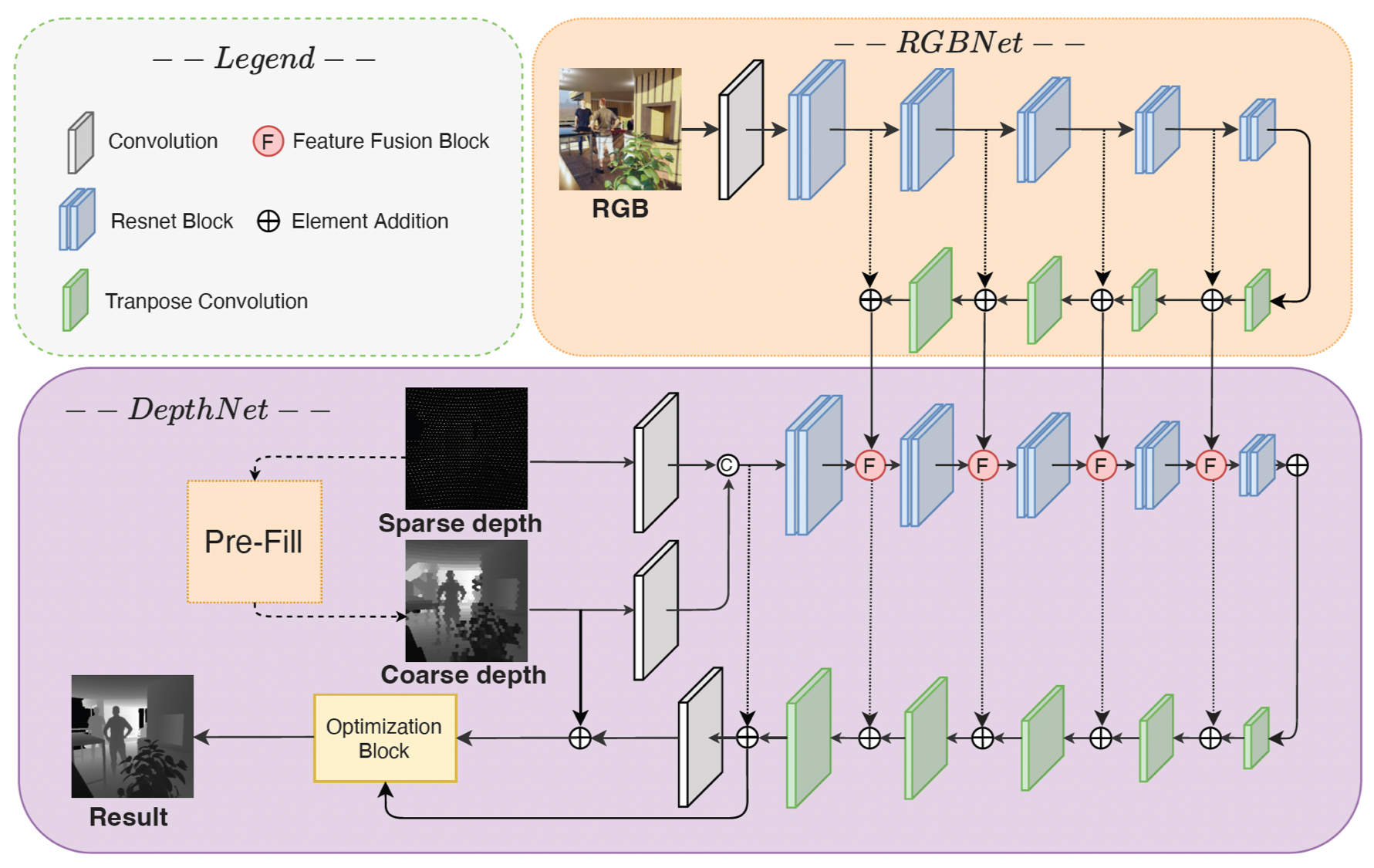}
\caption{Model architecture of NPU-CVR team.}
\label{fig:npu-cvr-network}
\end{figure}
Team NPU-CVR submitted two results with a difference in network structural parameter setting. 
To efficiently fill the sparse and noisy depth maps captured by the ToF camera into dense depth maps, they
propose \wenxiu{an} RGB-guided depth completion method. The overall framework of their method shown in Fig.~\ref{fig:npu-cvr-network} is based on residual learning. Given a sparse depth map, they first fill it into a coarse dense depth map by pre-processing.
Then, they obtain the residual by the network based on the proposed guided feature fusion block, and the
residual is added to the coarse depth map to obtain the fine depth map. In addition, they also propose a depth
map optimization block \wenxiu{that} enables deformable convolution to further improve the performance of the method. 

\subsection{JingAM}
Team JingAM improves on the FusionNet proposed in
the paper~\cite{van2019sparse}. This paper takes RGB map and sparse depth map as
input. The network structure is divided into global network and local
network. The global information is obtained through the global network.
In addition, the confidence map is used to combine the two inputs according to the uncertainty in the later fusion method.  On this basis, they added the skip-level structure and more bottlenecks, and canceled the guidance
map and replaced it with the global depth prediction. The loss function
weights of the prediction map, global map, and local map are 1, 0.2, and
0.1 respectively. In terms of data enhancement, the brightness,
saturation, and contrast were randomly adjusted from 0.1 to 0.8. The
sampling step size of the depth map is randomly adjusted between 5 and
12, and change the input size from fixed size to random size. The series
of strategies they adopted significantly improved RMSE and MAE compared
to the paper~\cite{van2019sparse}.

\subsection{Anonymous}
\begin{figure}[!ht]
\setlength{\abovecaptionskip}{0.cm}
\setlength{\belowcaptionskip}{-0.cm}
\centering
\includegraphics[width=0.8\textwidth]{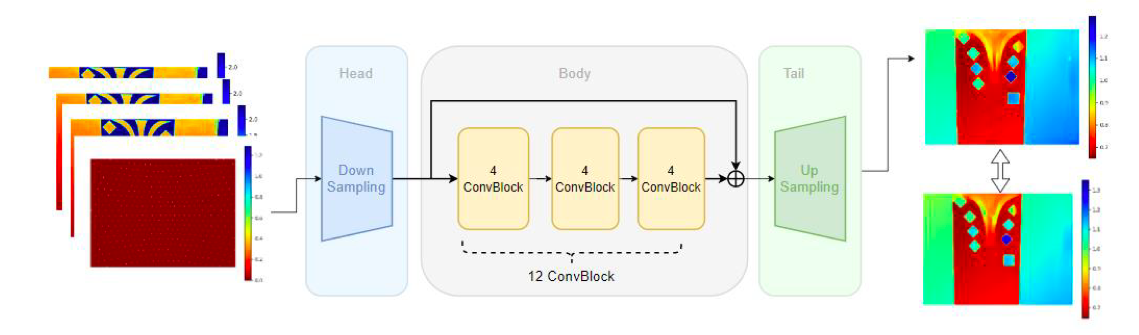}
\caption{Model architecture of Anonymous team.}
\label{fig:megvii-network}
\end{figure}
Team Anonymous proposed their method based on a fully convolutional neural network (FCN) which consists of 2 downsampling
convolutional layers, 12 residual blocks, and 2 upsampling convolutional layers to produce the final
result as shown in Fig.~\ref{fig:megvii-network}.  During the training, 25 sequential images and their sparse depth maps from one scene of
the provided training dataset will be selected as inputs, then the FCN will predict 25 dense depth maps.
The predicted dense depth maps and their ground truth will be used to calculate element-wise training
loss consisting of L1 loss, L2 loss, and RMAE loss. Furthermore, RTSD will be calculated with the
predicted dense depth maps and regarded as one item of the training loss. During the evaluation, the
input RGB image and the sparse depth map will also be concatenated along the channel \wenxiu{dimension}
and sent to the FCN to predict the final dense depth map.

\subsection{MainHouse113}
\begin{figure}[!ht]
\setlength{\abovecaptionskip}{0.cm}
\setlength{\belowcaptionskip}{-0.cm}
\centering
\includegraphics[width=0.8\textwidth]{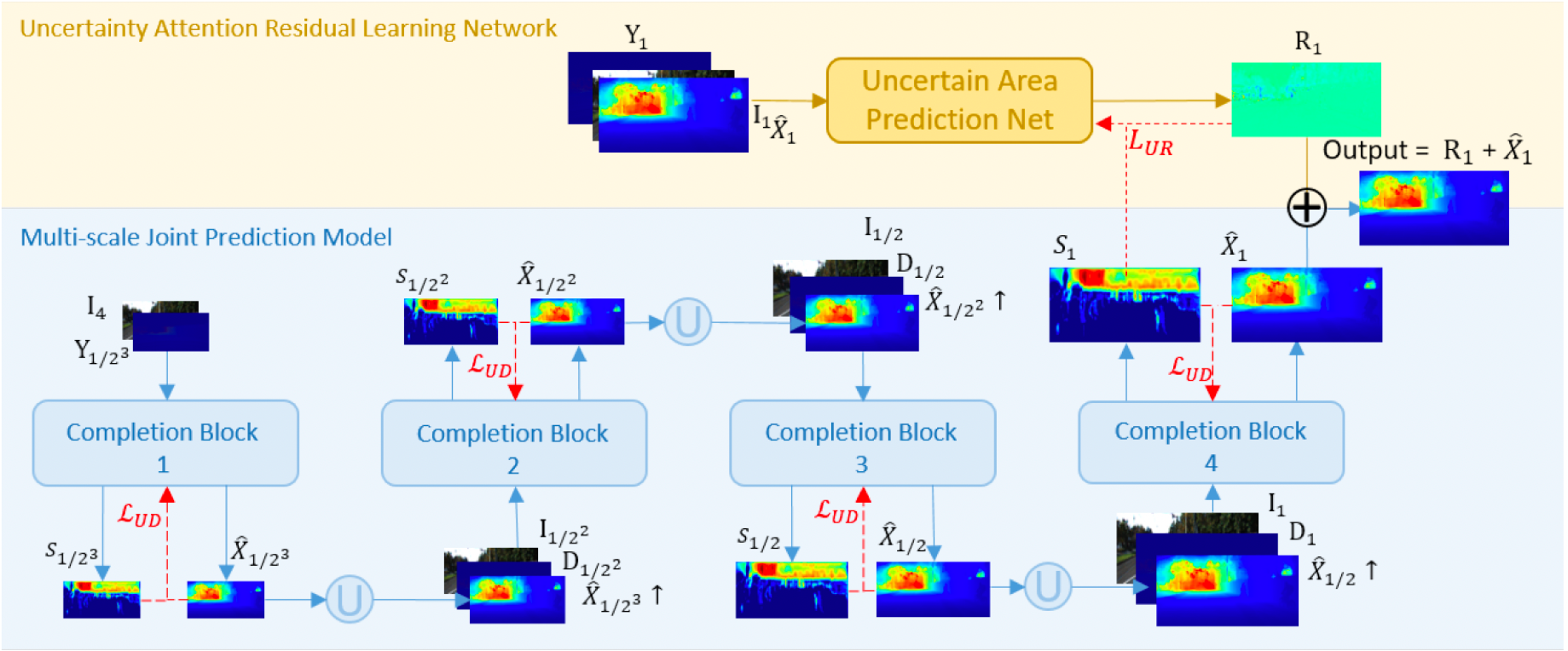}
\caption{Model architecture of Mainhouse113 team.}
\label{fig:mainhouse113-network}
\end{figure}
As shown in Fig.~\ref{fig:mainhouse113-network}, Team Mainhouse113 uses a multi-scale joint prediction network (MSPNet) that inputs RGB image and
sparse depth image, and simultaneously predicts the dense depth image and uncertainty map. They introduce
the uncertainty-driven loss to guide network training and  leverage a course-to-fine strategy to
train the model. Second, they use an uncertainty attention residual learning network (UARNet), which
inputs RGB image, sparse depth image and dense depth image from MSPNet, and outputs residual
dense depth image. The uncertainty map from MSPNet serves as an attention map when training the
UARNet. The final depth prediction is the element-sum of the two dense depth images from the two
networks above. To meet the speed requirement, they use depthwise separable convolution instead of
ordinary convolution in the first step, which may \wenxiu{lose} a few effects.
\subsection{UCLA Vision Lab}
\begin{figure}[!ht]
\setlength{\abovecaptionskip}{0.cm}
\setlength{\belowcaptionskip}{-0.cm}
\centering
\includegraphics[width=0.8\textwidth]{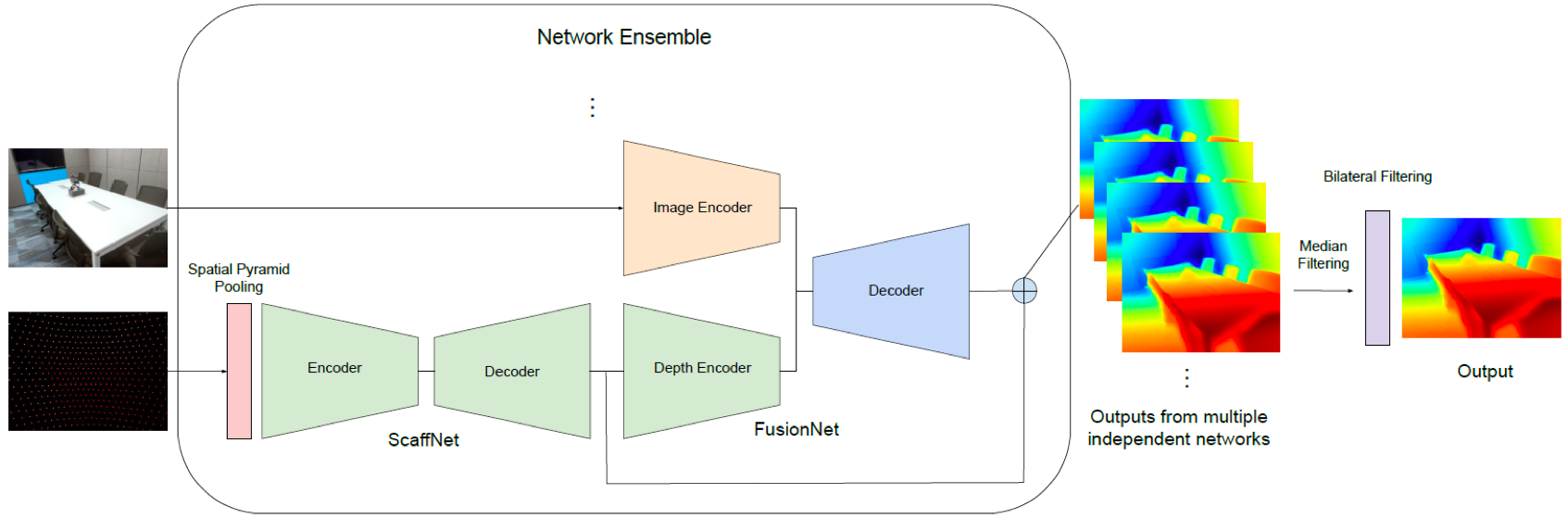}
\caption{Model architecture of UCLA team.}
\label{fig:ucla-network}
\end{figure}
Team UCLA Vision Lab proposed a method that aims to learn a prior on the shapes populating the scene from
only sparse points. This is realized as a light-weight encoder-decoder network (ScaffNet)~\cite{wong2021learning} and because
it is only conditioned on the sparse points, it generalizes across domains. A second network, also an
encoder-decoder network (FusionNet)~\cite{tang2020learning}, takes the putative depth map and the RGB colored image as
input using separate encoder branches and outputs the residual to refine the prior. Because the two
networks are extremely light-weight (13ms per forward pass), they are able to instantiate an ensemble
of them to yield robust predictions. Finally, they perform an image-guided bilateral filtering step on the
output depth map to smooth out any spurious predictions. The overall model architecture during inference time is shown in Fig.~\ref{fig:ucla-network}.

\section{Conclusions}
In this paper, we summarized the RGB+ToF Depth Completion challenge in the first Mobile Intelligent Photography and Imaging workshop (MIPI 2022) held in conjunction with ECCV 2022. The participants were provided with a high-quality training/testing dataset, which is now available for researchers to download for future \wenxiu{research}. We are excited to see the new progress contributed by the submitted solutions in such a short time, which are all described in this paper. The challenge results are reported and analyzed. For future works, there \wenxiu{is} still plenty room \wenxiu{for} improvements including dealing with depth outliers/noises, precise depth boundaries, high depth resolution, dark scenes, as well as low latency and low power consumption, etc.

\section{Acknowledgements}
We thank Shanghai Artificial Intelligence Laboratory, Sony, and Nanyang Technological University to \wenxiu{sponsor} this MIPI 2022 challenge. We thank all the organizers and all the participants for their great work. 

\appendix
\section{Teams and Affiliations}
\label{appendix:teams}
\begin{multicols}{2}       %
\scriptsize
\noindent\textbf{ZoomNeXt Team} \\
\textbf{Title}: Learning An Efficient Multimodal Depth Completion Model \\
\textbf{Members}:
$^{1}$Dewang Hou (dewh@pku.edu.cn), $^{2}$Kai Zhao \\
\textbf{Affiliations}:
$^{1}$Peking University, $^{2}$Tsinghua University
\\
\\
\textbf{GAMEON Team} \\
\textbf{Title}: A multi-scale depth completion network with high stability \\
\textbf{Members}:
$^{1}$Liying Lu (lylu@cse.cuhk.edu.hk), $^{2}$Yu Li, 
$^{1}$Huaijia Lin,
$^{3}$Ruizheng Wu,
$^{3}$Jiangbo Lu,
$^{1}$Jiaya Jia\\
\textbf{Affiliations}:
$^{1}$The Chinese University of Hong Kong, $^{2}$International Digital Economy Academy (IDEA),
$^{3}$SmartMore
\\
\\
\textbf{Singer Team}\\
\textbf{Title}: 
Depth Completion Using Laplacian Pyramid-Based Depth Residuals\\
\textbf{Members}:
Qiang Liu (476582539@qq.com), Haosong Yue, Danyang Cao, Lehang Yu, Jiaxuan Quan, Jixiang Liang\\
\textbf{Affiliations}:
BeiHang University
\\
\\
\textbf{NPU-CVR Team}\\
\textbf{Title}: 
An efficient residual network for depth completion of sparse
Time-of-Flight depth maps\\
\textbf{Members}:
Yufei Wang (wangyufei1951@gmail.com), Yuchao Dai, Peng Yang\\
\textbf{Affiliations}:
School of Electronics and Information, Northwestern Polytechnical University\\
\\
\\
\textbf{JingAM Team}\\
\textbf{Title}: 
Depth completion with RGB guidance and
confidence\\
\textbf{Members}:
Hu Yan (hu.yan@amlogic.com), Houbiao Liu, Siyuan Su, Xuanhe Li\\
\textbf{Affiliations}:
Amlogic, Shanghai, China\\
\\
\\
\textbf{Anonymous Team}\\
\textbf{Title}:
Prediction consistency is learning from yourself\\
\\
\\
\textbf{MainHouse113 Team}\\
\textbf{Title}:
Uncertainty-based deep learning framework with depthwise separable convolution for depth completion\\
\textbf{Members}:
Rui Ren (1019479834@qq.com), Yunlong Liu, Yufan Zhu\\
\textbf{Affiliations}:
Xidian University
\\
\\
\textbf{UCLA Vision Lab Team}\\
\textbf{Title}:
Learning shape priors from synthetic data for depth completion\\
\textbf{Members}:
$^{1}$Dong Lao (lao@cs.ucla.edu), Alex Wong, $^{1}$Katie Chang\\
\textbf{Affiliations}:
$^{1}$UCLA, $^{2}$Yale University
\\
\\
\end{multicols}

\clearpage
%
%
\bibliographystyle{splncs04}
\bibliography{egbib}
\end{document}